\documentclass[conference, 9pt, a4paper]{IEEEtran}
\IEEEoverridecommandlockouts
\usepackage{cite}
\usepackage{amsmath,amssymb,amsfonts}
\usepackage{algorithmic}
\usepackage{graphicx}
\usepackage{textcomp}
\usepackage{xcolor}
\usepackage{hyperref}
\usepackage{booktabs}
\usepackage{tabularx}
\usepackage{dcolumn} 
\usepackage{tikz}
\usepackage{csquotes}
\usepackage{ragged2e}
\usetikzlibrary{positioning, fit}

\usepackage{geometry}
\newcolumntype{d}[1]{D{.}{.}{#1}}

\def\BibTeX{{\rm B\kern-.05em{\sc i\kern-.025em b}\kern-.08em
    T\kern-.1667em\lower.7ex\hbox{E}\kern-.125emX}}
\begin{document}

\title{Classifying Tweet Sentiment Using the Hidden State and Attention Matrix of a Fine-tuned BERTweet Model}

\author{
    \IEEEauthorblockN{Tommaso Macrì}
    \IEEEauthorblockA{
        \textit{ETH Zürich}\\
        Zürich, Switzerland \\
        macrit@student.ethz.ch
    }
    \and
    \IEEEauthorblockN{Freya Murphy}
    \IEEEauthorblockA{
        \textit{ETH Zürich}\\
        Zürich, Switzerland \\
        fmurphy@student.ethz.ch
    }
    \and
    \IEEEauthorblockN{Yunfan Zou}
    \IEEEauthorblockA{
        \textit{ETH Zürich}\\
        Zürich, Switzerland \\
        yunzou@student.ethz.ch
    }
    \and
    \IEEEauthorblockN{Yves Zumbach}
    \IEEEauthorblockA{
        \textit{ETH Zürich}\\
        Zürich, Switzerland \\
        yzumbach@student.ethz.ch
    }
    \\
}

\maketitle

\begin{abstract}
This paper introduces a study on tweet sentiment classification.
Our task is to classify a tweet as either positive or negative.
We approach the problem in two steps, namely embedding and classifying.
Our baseline methods include several combinations of traditional embedding methods and classification algorithms. Furthermore, we explore the current state-of-the-art tweet analysis model, BERTweet, and propose a novel approach in which features are engineered from the hidden states and attention matrices of the model, inspired by empirical study of the tweets.
Using a multi-layer perceptron trained with a high dropout rate for classification, our proposed approach achieves a validation accuracy of 0.9111.
\end{abstract}

\section{Introduction}
Since its debut in 2006, Twitter has steadily grown in popularity and now has an estimated 199 million monthly active users\footnote{\url{https://www.businessofapps.com/data/twitter-statistics/}} posting a total of 500 million tweets per day\footnote{\url{https://blog.twitter.com/engineering/en_us/a/2013/new-tweets-per-second-record-and-how}}. This steady flow of data can be of interest when analyzing public opinion of political candidates, brands, events, etc. Hence there is an increasing interest in performing sentiment analysis on tweets to classify them as either positive or negative.

The nature of tweets poses several unique challenges from a natural language understanding perspective: tweets are typically short (limited to 280 characters) and make frequent use of slang, abbreviations, and hashtags. The informal language used in tweets often includes sarcasm and nuanced, context-specific references. This suggests that a sentiment classifier developed for newspaper articles for instance, may not perform as well on Twitter data as it does on its target corpus.

In this project, we study various sentiment analysis models for tweets. The general strategy is to first compute some embeddings to represent the information of the tweet, then run some classifiers on the embedding to decide whether the tweet has a positive or negative sentiment.

Apart from some baseline models with traditional embedding and classification algorithms, we make extensive use of the BERT\cite{BERT} (Bidirectional Encoder Representations from Transformers) for the task. More specifically, we adopt the BERTweet\cite{BERTweet} model, which is a variant of BERT pre-trained on tweets data, as the basis of our proposed model.
After fine-tuning the pre-trained BERTweet model on the given dataset, we investigate the effect of its hidden states and attention matrices.
From these data, we output a 1024 feature embedding of each tweet.
In turn, we use these embeddings to train a multi-layer perceptron with a high dropout rate.
Our final solution achieves a validation accuracy of 91.11\%.

\section{Models}

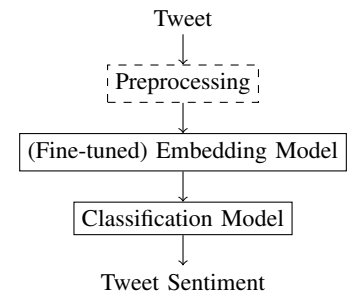
\begin{figure}[htbp]
\begin{center}
\begin{tikzpicture}[node distance=4mm]
	\node (tweet) {Tweet};
	\node[draw, dashed, below=of tweet] (preprocessing) {Preprocessing};
	\node[draw, below=of preprocessing] (embedding) {(Fine-tuned) Embedding Model};
	\node[draw, below=of embedding] (classif) {Classification Model};
	\node[below=of classif] (y) {Tweet Sentiment};
	
	\draw[->] (tweet) -- (preprocessing);
	\draw[->] (preprocessing) -- (embedding);
	\draw[->] (embedding) -- (classif);
	\draw[->] (classif) -- (y);
\end{tikzpicture}
\caption{\label{fig:model_pipeline}General pipeline used for the various classification models created during the project. Note that the dashed steps are not always required.}
\end{center}
\end{figure}

We give some high-level overview of the various embeddings and classification models that we used in this project.
The pipeline we used to produce tweet sentiment classification was always similar in nature and is detailed in Figure \ref{fig:model_pipeline}.

\subsection{Baseline Embeddings}

\textbf{Bag-of-words.}
Bag-of-words is a simple model where a text is represented by a multiset of its words. Only the frequency of each word is considered and word order is ignored. We use the bag-of-words model to extract the frequency of each word in a tweet and then use a classifier to classify the tweet as positive or negative.

\textbf{Word2vec.}
Word2vec was introduced by Mikolov et al. in 2013 \cite{word2vec1, word2vec2} as a model which also captures the similarity between words in a corpus.
This allows it to have some notion of the 'meaning' of a word or the relation that words have to each other.
It does this by representing each word as a vector where the cosine similarity between vectors indicates the similarity between the words represented by those vectors.

\textbf{Doc2vec.}
Doc2vec \cite{doc2vec} is a generalization of the word2vec embedding applied to entire documents or paragraphs rather than individual words.
In our case, we create a vector representation of each tweet and are thus able to infer the similarity of individual tweets to each other.

\textbf{GloVe.} GloVe \cite{GloVe} combines the local context window method used by word2vec with global matrix factorization to represent words as numerical vectors.

\subsection{BERT and BERTweet Embeddings}

BERT stands for Bidirectional Encoder Representations from Transformers.
It is a transformer-based model introduced in \cite{BERT}.

A transformer-based model is a type of deep learning model that uses attention mechanisms, i.e. it takes the context of the input into account.
Attention, in the context of deep learning, is a technique used to make the models (neural networks, generally speaking) identify the important part in the input and give less importance to the other parts.
Recognizing the important part is learned during the training process of the model.

The BERT model was trained by Google on a large corpus of English language data including Wikipedia.
Upon publication it achieved state-of-the-art performance for natural language processing tasks including GLUE (General Language Understanding Evaluation) \cite{GLUE}, SQuAD (Standford Question Answering Dataset) \cite{SQuAD}, and SWAG (Situations With Adversarial Generations) \cite{SWAG}.
Being based on transformers, this model is able to take the context of a word into account.
This means that it can differentiate the various meanings of a word, i.e. produce different embeddings, given its context.
Even though other transformer-based models existed before BERT (for example GPT Transformer model), BERT distinguished itself by being bi-directional: it takes the left \textit{and} right context of a word into account whereas previous models only took the left context into account.

BERTweet is a model following the BERT architecture but trained on a corpus of 850M English tweets. It achieves state-of-the-art performance in tasks such as part-of-speech-tagging, named-entity recognition, and text classification, including sentiment analysis \cite{BERTweet}.

In our novel approach, we use the BERTweet model to produce tweet embeddings that we extract from the model state after feeding it the tweets. 

\subsection{Classifiers}

\textbf{Random forest.}
In the random forest classification method, many decision trees with limited depth are trained on the data.
The actual outcome of the classification model is the majority class as produced by the various trees \cite{randomforest}.

\textbf{AdaBoost.}
Ensemble models aim to improve accuracy by combining the outcome of multiple base models, usually decision trees, to create a better classifier.
Boosting algorithms, like AdaBoost (Adaptive Boosting), adjust the weights of each submodel at each iteration to focus on difficult observations and thus improve classification accuracy \cite{adaboost}.

\textbf{Gradient boosting.}
Gradient boosting iteratively adds 'weak learners' (decision trees) to a decision forest using gradient descent to minimize the loss \cite{gradientboosting}.
This additive model gradually improves the base decision forest accuracy.

\textbf{Fully connected network.}
A fully connected neural network is a basic type of neural network where all the nodes in one layer are connected to all the nodes in the next layer.

\textbf{Simple LSTM.}
Long short-term memory (LSTM) is an architecture used with recurrent neural networks (RNNs) to solve the vanishing gradient problem \cite{lstm}.
This allows the RNN to better learn long-term dependencies in time series data.
In our case, using an LSTM model allows us to retain word sequence information.

\textbf{CNN-LSTM.}
Convolutional neural networks (CNNs) are particularly well-suited to classifying data with spatial structure such as such pixels in an image or words in a document. They work by repeatedly using a kernel to compute convolutions over data, thus reducing the dimensionality of the data while retaining key spatial information. A CNN can be used as an initial feature extraction layer for an LSTM model to further support word sequence prediction \cite{cnnlstm}.
 
\section{Novel Approach}

This section introduces the implementation of our final proposed methods.
The key steps include generating tweet embeddings using the BERTweet model and training a classifier on these embeddings. The former requires BERTweet-specific sentence pre-processing, fine-tuning of the BERTweet model and feature extraction, while the latter implies training a fully-connected neural network as a classifier of the tweet sentiment given our custom tweet embedding.

Our novel idea is to generate tweet embeddings that included not only parts of the hidden states of the BERTweet model but also parts of the attention matrix.
For each tweet, we generate 768 features from the hidden state layers and 256 features from the attention matrix.
The procedure to generate these features is detailed below.

\subsection{Sentence Pre-processing}

The raw training dataset includes 2.5 million tweets in the form of raw texts.
Before feeding inputs to the BERTweet model, they need to be tokenized, i.e. words are split into tokens then encoded as the sequence of token IDs.
To perform this task, we use the pre-trained BERTweet tokenizer.

\begin{figure}[!htbp]
\centering
\includegraphics[height=4cm, width=8cm]{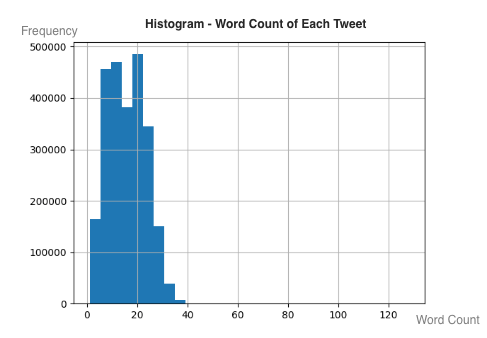}
\caption{\label{fig:histogram}Histogram of Word Count per Tweet}
\end{figure}

When generating the hidden layer features, we set each tweet's encoding to have a length of 50 (using padding or cropping depending on the tweet length) since most of the tweets have a length smaller than 50 words, as suggested by the histogram of the word counts for the provided tweet dataset in Figure \ref{fig:histogram}.
Note that the encoding of a tweet also contains an attention mask, i.e. a binary mask indicating which tokens are padding.

We remove this limitation when generating the attention map features for reasons that are detailed below.

\subsection{BERTweet Fine-tuning}

The pre-trained BERTweet model is not intended specifically for sentiment classification \cite{BERTweet}.
The model still computes data that can be used to perform this task, however, and using some transfer-learning methods, e.g. fine-tuning, BERTweet can yield much better results on the task.
Fine-tuning is the process of further training a pre-trained model either using task-specific data or changing the output layer to fit a specific task.

To fine-tune BERTweet for our task, we update the entire model parameters using the dataset of our task.
This process helps the model retain information related to the tweet sentiment classification task specifically.

Because the model is very large --- billions of parameters --- we only fine-tune it using 200K tweets from the complete dataset to improve performance.
We do 2 full epochs using the \texttt{AdamW} optimizer from the \texttt{transformers} library, using a learning rate of 1e-5 and a batch size of 128.

\subsection{Feature Engineering: Hidden States and Attention Matrices}

We now use the fine-tuned BERTweet model to produce tweet embeddings.
We use values from the model's hidden state and attention matrix.

Generally speaking, it is easier to work with fixed length embeddings.
This raises the following questions:

\begin{enumerate}
    \item What activation values should be extracted, i.e. what should the embedding of a single token be?
    \item How to merge or combine the embedding of each token in the tweet, i.e. how to produce a fixed-length embedding for each tweet?
\end{enumerate}

Regarding the first question, we select the features to use for the embedding empirically.
This means we try several embeddings using the partial dataset, i.e. last layer activation, last two layers activation, last four layers activation and sum of last 4 layers activation; and find out that the concatenation of the activation values of the last four layers of the model produce the best results. This is also indicated by other empirical experiments online \cite{illustrated-bert}.
Each token now produces 768 features.

Now we need to combine the embedding of each token into a fixed-length encoding for the tweet.
We use a simple approach: average all the token embeddings.
Note that averaging all the embeddings means losing a lot of data which could be useful for classifying the tweet's sentiment.
Some future work might look into how to use variable length embeddings or finding a better way than using the token's embeddings average.

We decided to keep the number of features low, using only 768 features per tweet, where each feature is encoded in 4 bytes and with 2.5M tweets since the hidden state features already represent 7.68GB of data.
Keeping each token embedding would mean multiplying the size of the dataset by 50.
Training models on more than 250GB of data is a challenge in itself and we deemed it to be out-of-scope for this project.
Furthermore, using this simple approach we are already able to beat all of our baselines.

We now produce the 256 attention matrix features.
The attention matrix in the BERTweet model encodes how much attention should be given to words in the context of another word to infer its meaning.
The intuition is that some words in the context of a word are more important to infer its meaning than others.
We hypothesize that using the attention matrix can improve the accuracy of the classifier.

\begin{figure}[!htbp]
\centering
\includegraphics[height=3.5cm, width=8cm]{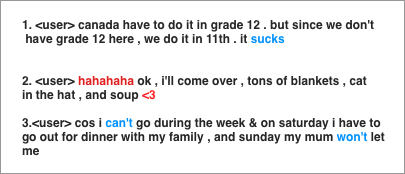}
\caption{\label{fig:example_sentences}Examples of tweets with emotional words (red for positive, blue for negative) at the beginning and end}
\end{figure}

For a tweet that is encoded into $n$ tokens, the attention matrix will be $n\times n$.
To turn this into a fixed-length feature vector, we first have a look at the tweets.
As illustrated in Figure \ref{fig:example_sentences}, we remark that words at the beginning and end of the tweets are often highly relevant to predict a tweet's sentiment.

For this reason, we decide to keep the attention matrix values that concern the first and last tokens of the tweet encoding.
More specifically, we only keep the values of the four $8\times 8$ submatrices located at the corners of the attention matrix as illustrated in Figure \ref{fig:attention_matrix}.
This produces a feature vector of length 256.

\begin{figure}[ht!]
    \centering
    \newcommand{\tikzmark}[2]{\tikz[overlay,remember picture,baseline=(#1.base)] \node (#1) {$#2$};}
	\[
	A = \left(
	\begin{array}{*7{c}}
		\tikzmark{one}{a_{1,1}} & \cdots & a_{1,\frac{k}{2}} & \cdots & \tikzmark{three}{a_{1,n-\frac{k}{2}}} & \cdots & a_{1,n}\\
		a_{2,1} & \cdots & \tikzmark{two}{a_{2,\frac{k}{2}}} & \cdots & a_{2,n-\frac{k}{2}} & \cdots & \tikzmark{four}{a_{2,n}}\\
		\vdots & \vdots & \vdots & \ddots & \vdots & \vdots & \vdots\\
		\tikzmark{five}{a_{n-1,1}} & \cdots & a_{n-1,\frac{k}{2}} & \cdots & \tikzmark{seven}{a_{n-1,n-\frac{k}{2}}} & \cdots & a_{n-1,n}\\
		a_{n,1} & \cdots & \tikzmark{six}{a_{n,\frac{k}{2}}} & \cdots & a_{n,n-\frac{k}{2}} & \cdots & \tikzmark{eight}{a_{n, n}}\\
	\end{array}
	\right)
	\]
	
	\tikzset{%
		highlight/.style={
			rectangle,
			fill=red!15,
			fill opacity=0.4,
			inner sep=0pt}
	}
	\begin{tikzpicture}[remember picture, overlay, node distance=0mm]
		\node[fit=(one.north west) (two.south east), highlight] (box_one) {};
		\node[fit=(three.north west) (four.south east), highlight] (box_two) {};
		\node[fit=(five.north west) (six.south east), highlight] (box_three) {};
		\node[fit=(seven.north west) (eight.south east), highlight] (box_four) {};
		
		\node[above=of box_one] {$k/2 \times k/2$};
	\end{tikzpicture}
	
    \caption{Values extracted from the attention matrix in order to produce fixed-length features when $n > k$ where $A$ is $n\times n$.}
    \label{fig:attention_matrix}
\end{figure}

In total, we now have 768 features from the hidden state and 256 features from the attention matrix yielding an embedding of length 1024 for each tweet.

\subsection{Multi-layer Perceptron Classifier}

From the tweet embedding, we want to produce a tweet sentiment classification.
To this end, we use a fully-connected, multi-layer perceptron classifier.
Non-linearity is provided by feeding the output of each layer through ReLU functions.

We use works from Hanin \cite{fully_connected} to define the width the layers of our neural network.
He suggests that the width of each layer in a ReLU neural network should be slightly higher than the number of features to be able to approximate functions well under some regularity conditions.
Our embeddings have 1024 features, therefore we use layers with 1500 neurons.

We train our MLP with a batch size of 16384 and a learning rate of 5e-5 using the Adam optimizer for 15 epochs and the binary cross-entropy loss function.

To avoid overfitting, we use several further strategies detailed below.

\subsubsection{Small Neural Network Depth}

A simple yet very effective way of preventing overfitting is to use models that are not too deep.
For this task we use 3 hidden layers, each containing 1500 neurons.
Empirically, this proves sufficient to approximate the sentiment function with high enough accuracy.

\subsubsection{High Dropout Rate}

A second method to prevent overfitting when training a model is to use a method call \enquote{dropout}.
When training the model with a dropout rate $p$, each forward pass ignores a fraction $p$ of the neurons randomly and re-scales the output by $\frac{1}{1-p}$.
The dropped parameters are not updated in the backward pass.

In our case, we set $p$ to a large value: $p = 0.9$.
This is similar to an ensemble of light-weight models with much fewer parameters.
A high dropout rate efficiently prevents the model from overfitting the dataset.
During the training, using a high dropout rate increases our validation accuracy and training accuracy steadily.

\subsection{Majority Voting}

\begin{figure}[!htbp]
\centering
\includegraphics[height=5cm, width=8cm]{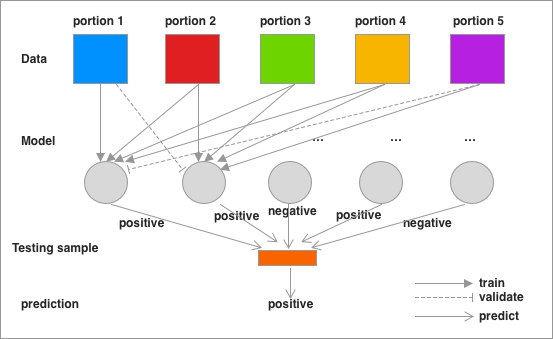}
\caption{\label{fig:cv_major_vote}5-Fold Majority Voting}
\end{figure}

Our final classifier was an ensemble model containing 5 MLPs performing majority voting.
Instead of training only one classifier on all the data, we split the data in 5 folds and train 5 models, dropping one of the folds when training each of the models.
We then perform majority voting on the five classifiers to get the final tweet sentiment.
This process is summarized in Figure \ref{fig:cv_major_vote}.

The reasons for doing this were mainly related to computational limitations: we could not load the entire training dataset because of memory limitations.
Nevertheless, doing this kind of resampling and aggregation, akin to bootstrapping, can also improve accuracy and prevent overfitting.

\subsection{The Case for Not Using BERTweet to Produce the Classification Directly}

We remark that the BERTweet model can be fine-tuned to produce tweet sentiment directly, without using an additional classifier model.
We tried this strategy but accuracy was lower than using our custom tweet embedding and classifiers.
A possible explanation is that the neuron which performs the classification only has access to the activation of the last layer whereas our embeddings contained the activations of the four last layers.
Further, the information of the attention matrix is also lacking.

\section{Other Attempts -- Text Expansion}
\label{sec:text_expansion}

We wanted to explore another idea in the context of this project, namely to expand each tweet by feeding it to a generative model like GPT-3 from OpenAI \cite{gpt3} or GPT-J \cite{gpt-j}.
From a single tweet of 240 characters at most, we wanted to generate a longer text.
Our hopes were that the GPT model would be able to understand the sentiment in the tweet and produce text with the same sentiment.
The classification task would then become easier as we would have had more text to do the classification on.
The core insight is that the GPT model, being very large, may better understand the sentiment of the tweet and ease the task of the classifier.

We were not able to materialize this idea because of performance limitations:
\begin{description}
    \item[GPT-3] is a proprietary model to which we do not have access.
    \item[GPT-J] is open source but takes approximately 1 second to extend a tweet.
    With a dataset of 2.5 million tweets, it would have taken multiple weeks to produce the text for every tweet, which we deemed unreasonable.
\end{description}

\section{Results}

The validation accuracy and public score on the Kaggle leaderboard that we were able to reach using various baselines and our novel models are given in Table \ref{tab:results}.

\begin{table}[htbp]
\centering
\caption{Accuracies and leaderboard score of various models.}
\label{tab:results}
\begin{tabularx}{\linewidth}{>{\RaggedRight}X d{-1} l}
    \toprule
    \textbf{Model} & \multicolumn{1}{r}{\textbf{Accuracy}} & \multicolumn{1}{r}{\textbf{Leaderboard}} \\
    \midrule
    Bag-of-words + logistic regression & 0.8073\\
    Bag-of-words + random forest & 0.8121 \\
    Bag-of-words + AdaBoost & 0.7065 \\
    Bag-of-words + gradient boosting & 0.7292 \\
    \midrule
    Word2vec + MLP & 0.7769 \\
    Word2vec + simple LSTM & 0.6716 \\
    Word2vec + CNN-LSTM & 0.6937 \\
    \midrule
    Doc2vec + logistic regression & 0.6784 \\
    Doc2vec + random forest & 0.6968 \\
    Doc2vec + AdaBoost & 0.6404 \\
    Doc2vec + gradient boosting & 0.6630 \\
    \midrule
    GloVe + logistic regression & 0.7410 \\
    GloVe + random forest & 0.7554 \\
    GloVe + AdaBoost & 0.6906 \\
    GloVe + gradient boosting & 0.7322 \\
    GloVe + XGBoost & 0.7843 \\
    \midrule
    BERT + logistic regression & 0.8320 \\
    BERT + random forest & 0.8108 \\
    BERT + AdaBoost & 0.7707 \\
    BERT + gradient boosting & 0.8007 \\
    BERT + XGBoost & 0.8318 \\
    \midrule
    BERTweet + logistic regression & 0.8801 \\
    BERTweet + random forest & 0.8578 \\
    BERTweet + AdaBoost & 0.8183 \\
    BERTweet + gradient boosting & 0.8489 \\
    BERTweet + XGBoost & 0.8771 \\
    BERTweet Built-in & 0.4258 \\
    BERTweet + Hidden Features + MLP & 0.8960 & 0.8880\\
    BERTweet + Hidden Features + MLP (Majority Vote) & & 0.8868\\
    \midrule
    BERTweet (fine-tuned) Built-in & 0.9081 & 0.9008\\
    BERTweet (fine-tuned) + Hidden Features + MLP & 0.9109 & 0.9018\\
    BERTweet (fine-tuned) + Hidden Features + MLP (Majority Vote) &  & 0.9018 \\
    BERTweet (fine-tuned) + Hidden Features + Attention Matrices + MLP & 0.9111 & 0.9024 \\
    BERTweet (fine-tuned) + Hidden Features + Attention Matrices + MLP (Majority Vote) & & \textbf{0.9034} \\
    \bottomrule
\end{tabularx}
\end{table}

\section{Discussion}
According to the results in Table \ref{tab:results}, BERT-based approaches perform better than the simpler baselines. The BERT models are very deep (billions of parameters), thus they are able to develop a better understanding of human languages than the simpler models used for our baselines.

That BERTweet works better than BERT on tweets is not a surprise since the language used on Twitter is very different from that used on Wikipedia.
Tweets often contain slang, abbreviations, typos, and hashtags, which are uncommon on Wikipedia. Thus a model trained to recognize these features produces better results.

Fine-tuning the BERTweet model improves the validation accuracy by around 1.5\% compared to the best non-fine-tuned model in the worst case.
This highlights the importance of fine-tuning the BERTweet model to the task at hand.

Our proposed custom embedding strategy outperforms the built-in classifier of the fine-tuned BERTweet model.
Using the activation of the last four layers instead of the activation of the last layer also proves to be very powerful at improving the accuracy (improvements of 1\% and 0.1\% on the validation set accuracy and leaderboard accuracy respectively).
We also remark that the data from the attention matrix improves the accuracy by around 0.6\% which seems to indicate that it provides relevant information to the classifier.

We postulate that our custom embedding works well because it keeps some lower-level information, i.e. we keep data from multiple layers, not just the last one, and, by including the attention matrix, it does not lose contextual and ordering information compared to the approach that only takes the averaged hidden states into consideration.

\section{Future Improvements}

In the future, we might be interested in running BERTweet on expanded versions of the tweets as described in Section \ref{sec:text_expansion}.
The tweets could be extended by asking generative models like GPT-J to produce more text showcasing the same sentiment as the original tweet.

Some alternatives could be explored to averaging the token embeddings in order to preserve the contextual and ordering information more efficiently than through the attention matrix.
Alternatively selecting some other features from the attention matrix could prove an interesting method.

\end{document}